\documentclass{article}

\usepackage[final,nonatbib]{nips_2017}

\usepackage[utf8]{inputenc} % allow utf-8 input
\usepackage[T1]{fontenc}    % use 8-bit T1 fonts
\usepackage{hyperref}       % hyperlinks
\usepackage{url}            % simple URL typesetting
\usepackage{booktabs}       % professional-quality tables
\usepackage{amsfonts}       % blackboard math symbols
\usepackage{nicefrac}       % compact symbols for 1/2, etc.
\usepackage{microtype}      % microtypography

\usepackage{soul}
\usepackage{subcaption}
\usepackage{graphicx}
\usepackage{wrapfig}
\usepackage{amsmath}
\usepackage{enumitem}

\usepackage{color}

\newcommand{\xx}{\mathbf{x}}
\newcommand{\lll}{\mathbf{l}} % label sequence

\title{Analyzing Hidden Representations in End-to-End Automatic Speech Recognition Systems}

\author{
  Yonatan Belinkov  \textnormal{and}  James Glass \\
  Computer Science and Artificial Intelligence Laboratory \\
  Massachusetts Institute of Technology \\
  Cambridge, MA 02139 \\
  \texttt{\{belinkov, glass\}@mit.edu} \\
}

\begin{document}

\maketitle

\begin{abstract}
Neural %network 
models have become ubiquitous in automatic speech recognition systems. While neural networks are typically used as acoustic models in more complex  systems, recent studies have explored end-to-end speech recognition systems based on neural networks, which can be trained to directly predict text from input acoustic features. 
Although such systems are conceptually elegant and simpler than traditional systems, it is less obvious how to interpret the trained models. 

In this work, we analyze the speech representations learned by a deep end-to-end model that is based on convolutional and recurrent layers, and trained with a connectionist temporal classification (CTC) loss. We use a pre-trained model to generate frame-level features which are given to a classifier that is trained on frame classification into phones. 
We evaluate representations from different layers of the deep model and compare their quality for predicting phone labels. 
Our experiments shed light on important aspects of the end-to-end model such as layer depth, model complexity, and other design choices. 
\end{abstract}

\section{Introduction}
Traditional automatic speech recognition (ASR) systems are composed of multiple components, including an acoustic model, a language model, a lexicon, and possibly other components. Each of these is trained independently and combined during decoding. As such, the system is not directly trained on the speech recognition task from start to end. 
In contrast, end-to-end ASR systems aim to map acoustic features directly to text (words or characters). Such 
models have recently become popular in the ASR community thanks to their simple and elegant architecture~\cite{icml2014c2_graves14,miao2015eesen,chorowski2014end,chan2015listen}.  
Given sufficient training data, they also perform fairly well. 
Importantly, such models do not receive explicit phonetic supervision, in contrast to traditional systems that typically rely on an acoustic model trained to predict phonetic units (e.g.\ HMM phone states). Intuitively, though, end-to-end models have to generate some internal representation that allows them to abstract over phonological units. For instance, a model that needs to generate the word ``bought'' should learn that in this case ``g'' is not pronounced as the phoneme /g/. 

In this work, we investigate if and to what extent end-to-end models \textit{implicitly} learn phonetic representations. The hypothesis is that such models need to create and exploit internal representations that correspond to phonetic units in order to perform well on the speech recognition task. Given a pre-trained end-to-end ASR system, we use it to extract frame-level features
from an acoustic signal. 
For example, these may be the hidden representations of a recurrent neural network (RNN) in the end-to-end system. We then feed these features to a classifier that is trained to predict a phonetic property of interest such as phone recognition. Finally, we evaluate the performance of the classifier as a measure of the quality of the input features, and by proxy the quality of the original %end-to-end
ASR system. 

We aim to provide quantitative answers to the following questions:
\begin{enumerate}[topsep=0pt]
\item To what extent do end-to-end ASR systems learn phonetic information?
\item Which components of the system capture more phonetic information? 
\item Do more complicated models learn better representations for phonology? And is ASR performance correlated with the quality of the learned representations? 
\end{enumerate}

Two main types of end-to-end models for speech recognition have been proposed in the literature: connectionist temporal classification (CTC) \cite{icml2014c2_graves14,miao2015eesen} and sequence-to-sequence learning (seq2seq) \cite{chorowski2014end,chan2015listen}. 
We focus here on CTC and leave exploration of the seq2seq model for future work.

We use a phoneme-segmented dataset for the phoneme recognition task, as it comes with time segmentation, which allows for accurate mapping between speech frames and phone labels. 
We define a frame classification task, where given representations from the CTC model, we  need to classify each frame into a corresponding phone label. 
More complicated tasks can be conceived of---for example predicting a single phone given all of its aligned frames---but  classifying frames is a basic and important task to start with.

Our experiments reveal that the lowest layers in a deep end-to-end model are best suited for representing phonetic information. 
Applying one convolution on input features improves the representation, but a  second convolution greatly degrades phone classification accuracy. 
Subsequent recurrent layers initially improve the quality of the representations. However, after a certain recurrent layer performance again drops, indicating that the top layers do not preserve all the phonetic information coming from the bottom layers. 
Finally, we cluster frame representations from different layers in the deep model and visualize them in 2D, observing different quality of grouping in different layers.

We hope that our results would promote the development of better ASR systems. For example, understanding representation learning at different layers of the end-to-end model can guide joint learning of phoneme recognition and ASR, as recently proposed in a multi-task learning framework~\cite{toshniwal2017multitask}. 

\vspace{-3pt}
\section{Related Work} \label{sec:related-work}
\vspace{-2pt}
\subsection{End-to-end ASR}
End-to-end models for ASR have become increasingly popular in recent years. Important studies include models based on 
connectionist temporal classification (CTC)~\cite{icml2014c2_graves14,miao2015eesen,Eyben:2009,amodei2016deep} and attention-based sequence-to-sequence models~\cite{chorowski2014end,chan2015listen,bahdanau2016end}. The CTC model is based on a recurrent neural network that takes as input acoustic features and is trained to predict a symbol per each frame. Symbols are typically characters, in addition to a special blank symbol. The CTC loss then marginalizes over all possible sequences of symbols given a transcription. 
The sequence-to-sequence approach, on the other hand, first encodes the sequence of acoustic features into a single vector and then decodes that vector into the sequence of symbols (characters). The attention mechanism improves upon this method by conditioning on a different summary of the input sequence at each decoding step. 

Both these of these approaches to end-to-end ASR usually predict a sequence of characters, although there have also been initial attempts at directly predicting words~\cite{soltau2016neural,audhkhasi2017direct}. 

\subsection{Analysis of neural representations}
While end-to-end neural network models offer an elegant and relatively simple architecture, they are often thought to be opaque and uninterpretable. Thus researchers have started investigating what such models learn during the training process. For instance, previous work evaluated neural network acoustic models on phoneme recognition using different acoustic features~\cite{mohamed2012understanding} or investigated how such models learn invariant representations \cite{yu2013feature} and encode linguistic features \cite{nagamine2015exploring,Nagamine+2016}. Others have correlated activations of gated recurrent networks with phoneme boundaries in autoencoders~\cite{wang2017gate} and in a text-to-speech system~\cite{wu2016investigating}.
A joint audio-visual model of speech and lip movements was developed in~\cite{chaabouni2017learning}, where phoneme embeddings were shown to be closer to certain linguistic features than embeddings based on audio alone.  Other joint audio-visual models have also analyzed the learned representations in different ways~\cite{chrupala2017representations,harwath2017learning,K17-1037}. 
Finally, we note that analyzing neural representations has also attracted attention in other domains like vision and natural language processing, including  machine translation~\cite{shi-padhi-knight:2016:EMNLP2016,belinkov:2017:ACL} and joint vision-language models~\cite{gelderloos-chrupala:2016:COLING}, among others. % studies. 
To our knowledge, hidden representations in end-to-end ASR systems have not been thoroughly analyzed before.

\section{Methodology}
We follow the following procedure for evaluating representations in end-to-end ASR models. First, we train an ASR system on a corpus of transcribed speech and freeze its parameters. Then, we use the pre-trained ASR model to extract frame-level feature representations on a phonemically transcribed corpus. Finally, we train a supervised classifier using the features coming from the ASR system, and evaluate classification performance on a held-out set. In this manner, we can obtain a quantitative measure of the quality of the representations that were learned by the end-to-end ASR model. A similar procedure has been previously applied to analyze a DNN-HMM phoneme recognition system \cite{Nagamine+2016} as well as text representations in neural machine translation models~\cite{shi-padhi-knight:2016:EMNLP2016,belinkov:2017:ACL}. 

More formally, 
let $\xx$ denote a sequence of acoustic features 
such as a spectrogram of frequency magnitudes. 
Let $\texttt{ASR}_t(\xx)$ 
denote the output of the ASR model at the $t$-th input.
Given a corresponding label sequence, $\lll$, 
we feed  $\texttt{ASR}_t(\xx)$ to a supervised classifier that is trained to predict a corresponding label, $l_t$. In the simplest case, we  have a label at each frame and perform frame classification. 
As we are interested in analyzing different components of the ASR model, we also extract features from different layers $k$, such that $\texttt{ASR}^{k}_t(\xx)$ denotes the output of the $k$-th layer at the $t$-th input frame.

We next describe the ASR model and the supervised classifier in more detail.

\subsection{ASR model} \label{sec:asr}

\begin{table}[t]
\centering
\caption{The ASR models used in this work.}
\label{tab:asr-models}
\begin{subtable}[t]{0.45\textwidth}
\caption{DeepSpeech2.}
\label{tab:deepspeech2}
\begin{tabular}{llrr}
\toprule
Layer & Type & Input Size & Output Size \\
\midrule
1 & cnn1 & 161 & 1952 \\
2 & cnn2 & 1952 & 1312 \\
3 & rnn1 & 1312 & 1760 \\
4 & rnn2 & 1760 & 1760 \\
5 & rnn3 & 1760 & 1760 \\
6 & rnn4 & 1760 & 1760 \\
7 & rnn5 & 1760 & 1760 \\
8 & rnn6 & 1760 & 1760 \\
9 & rnn7 & 1760 & 1760 \\ 
10 & fc & 1760 & 29 \\ 
\bottomrule
\end{tabular}
\end{subtable}
\hfill
\begin{subtable}[t]{0.45\textwidth}
\caption{DeepSpeech2-light.}
\label{tab:deepspeech2-light}
\begin{tabular}{llrr}
\toprule
Layer & Type & Input Size & Output Size \\
\midrule
1 & cnn1 & 161 & 1952 \\
2 & cnn2 & 1952 & 1312 \\
3 & lstm1 & 1312 & 600 \\
4 & lstm2 & 600 & 600 \\
5 & lstm3 & 600 & 600 \\
6 & lstm4 & 600 & 600 \\
7 & lstm5 & 600 & 600 \\
8 & fc & 600 & 29 \\ 
\bottomrule
\end{tabular}
\end{subtable}
%\vspace{-10pt}
\end{table}

The end-to-end model we use in this work is DeepSpeech2~\cite{amodei2016deep}, an acoustics-to-characters system based on a deep neural network. The input to the model is a sequence of audio spectrograms (frequency magnitudes), obtained with a 20ms Hamming window and a stride of 10ms. With a sampling rate of 16kHz, we have 161 dimensional input features.  Table~\ref{tab:deepspeech2} details the different layers in this model. 
The first two layers are convolutions where the number of output feature maps is 32 at each layer. The kernel sizes of the first and second convolutional layers are 41x11 and 21x11 respectively, where a convolution of TxF has a size T in the time domain and F in the frequency domain. Both convolutional layers have a stride of 2 in the time domain while the first layer also has a stride of 2 in the frequency domain. This setting results in 1952/1312 features per time frame after the first/second convolutional layers.

The convolutional layers are followed by 7 bidirectional recurrent layers, each with 
a hidden state size of 1760 dimensions. Notably, these are simple RNNs and not gated units such as long short-term memory networks (LSTM)~\cite{hochreiter1997long}, as this was found to produce better performance. We also consider a simpler version of the model, called DeepSpeech2-light, which has 5 layers of bidirectional LSTMs, each with 
600 dimensions (Table~\ref{tab:deepspeech2-light}). This model runs faster but leads to worse recognition results. %on LibriSpeech. 

Each convolutional or recurrent layer is followed by batch normalization~\cite{ioffe2015batch,laurent2016batch} and a ReLU non-linearity.  
The final layer is a fully-connected layer that maps onto the number of symbols (\mbox{29 symbols:}  26 English letters plus space, apostrophe, and a blank symbol). 

The network is trained with a CTC loss~\cite{graves2006connectionist}:
\begin{equation*}
L = - \log p(\lll|\xx)
\end{equation*}
where the probability of a label sequence $\lll$ given an input sequence $\xx$ is defined as:
\begin{equation*}
p(\lll|\xx) = \sum_{\pi \in \mathcal{B}^{-1}(\lll)} p(\pi | \xx) =  \sum_{\pi \in \mathcal{B}^{-1}(\lll)} \prod_{t=1}^T  \texttt{ASR}^{K}_t(\xx)[\pi_t]
\end{equation*}
where $ \mathcal{B}$ removes blanks and repeated symbols, 
$ \mathcal{B}^{-1}$ is its inverse image,
$T$ is the length of the label sequence $\lll$, and  $\texttt{ASR}^{K}_{t}(\xx)[j]$ is unit $j$ of the model output after the top softmax layer at time $t$, %which is 
interpreted as the probability of observing label $j$ at time $t$.   
This formulation allows mapping long frame sequences to short character sequences by marginalizing over all possible sequences containing blanks and duplicates.

\subsection{Supervised Classifier} \label{sec:classifier}
The frame classifier takes features from different layers of the DeepSpeech2 model as input  %
and predicts a phone label. 
The size of the input to the classifier thus depends on which layer in DeepSpeech2 is used to generate features. We model the classifier as a feed-forward neural network with one hidden layer, where the size of the hidden layer is set to 500.\footnote{We also experimented with a linear classifier and found that it produces lower results overall but leads to similar trends when comparing features from different layers.} 
This is followed by dropout (rate of 0.5) and a ReLU non-linearity, then a softmax layer mapping onto 
the label set size (the number of unique phones). 
We chose this simple formulation as we are interested in evaluating the quality of the representations learned by the ASR model, rather than improving the state-of-the-art on the supervised task.

We train the classifier with Adam~\cite{kingma2014adam} with the recommended parameters ($\alpha=0.001$, $\beta_1=0.9$, $\beta_2=0.999$, $\epsilon=e^{-8}$) to minimize the cross-entropy loss. We use a batch size of 16, 
train the model for 30 epochs, and choose the model with the best development loss for evaluation.

\section{Tools and Data} \label{sec:tools-data}
We use the 
\texttt{deepspeech.torch}~\cite{Naren2016}
 implementation of Baidu's DeepSpeech2 model~\cite{amodei2016deep}, which comes with  pre-trained models of both DeepSpeech2 and the simpler variant DeepSpeech2-light.  
 The end-to-end models are trained on LibriSpeech~\cite{panayotov2015librispeech}, a publicly available corpus of English read speech, containing 1,000 hours sampled at 16kHz.
The word error rates (WER) of the DeepSpeech2 and DeepSpeech2-light models on the Librispeech-test-clean dataset are 12 and 15, respectively, %~\cite{Naren2016}. 
 as reported in \cite{Naren2016}.  %

%  \begin{table*}[t]
% \centering
% \caption{Number of frames per phone in the TIMIT training set.}
% \label{tab:phone-frame-count}
% \begin{tabular}{lr|lr|lr|lr|lr}
% \toprule
% Phone & Frames & Phone & Frames & Phone & Frames & Phone & Frames & Phone & Frames \\
% \midrule
% s	&	69903	&	l	&	27095	&	axr	&	18872	&	el	&	8609	&	oy	&	5110	\\
% iy	&	44107	&	r	&	26315	&	pcl	&	18432	&	dh	&	8538	&	hv	&	5098	\\
% ix	&	37676	&	kcl	&	24610	&	ax	&	17060	&	ng	&	7279	&	en	&	4903	\\
% n	&	34397	&	ao	&	22937	&	pau	&	16711	&	gcl	&	7099	&	epi	&	3831	\\
% ih	&	33320	&	f	&	22801	&	q	&	16653	&	ch	&	7077	&	uh	&	3825	\\
% ae	&	31136	&	m	&	22250	&	sh	&	15571	&	th	&	6830	&	b	&	3770	\\
% z	&	30832	&	ow	&	21179	&	ux	&	14026	&	jh	&	6200	&	g	&	3179	\\
% eh	&	30533	&	er	&	20286	&	w	&	13393	&	hh	&	6059	&	nx	&	1940	\\
% ay	&	29922	&	ah	&	20136	&	v	&	12028	&	d	&	5875	&	zh	&	1217	\\
% tcl	&	29916	&	dcl	&	19995	&	bcl	&	11965	&	uw	&	5565	&	ax-h	&	1188	\\
% ey	&	28743	&	k	&	19790	&	aw	&	11704	&	y	&	5466	&	em	&	1010	\\
% aa	&	27753	&	t	&	19277	&	p	&	11455	&	dx	&	5367	&	eng	&	198	\\
% \bottomrule
% \end{tabular}
% \end{table*}

For the phoneme recognition task, we use TIMIT, which comes with 
time segmentation of phones. 
We use the official train/development/test split and extract frames for the frame classification task.
Table~\ref{tab:data-frame} summarizes statistics of the %extracted 
frame classification dataset. Note that due to sub-sampling at the DeepSpeech2 convolutional layers, the number of frames decreases by a factor of two after each convolutional layer. The possible labels are the 60 phone 
symbols included in TIMIT (excluding the begin/end silence 
symbol \textit{h\#}). 
%Table~\ref{tab:phone-frame-count} shows the number of frames per phone in the training set. 
%\begin{quote}
%\textit{
%aa	ae	ah	ao	aw	ax	axr	ay	b	bcl	ch	d	dcl	dh	dx	eh	el	em	en	eng	epi	er	ey	f	g	gcl	h	hh	hv	ih	ix	iy	jh	k	kcl	l	m	n	ng	nx	ow	oy	p	pau	pcl	q	r	s	sh	t	tcl	th	uh	uw	ux	v	w	y	z	zh
%}
%\end{quote}
We also experimented with the reduced set of 48 phones used by \cite{Lee-Hon:1989}.

\begin{table}[h]
\centering
\caption{Frame classification data extracted from TIMIT.}
\label{tab:data-frame}
\begin{tabular}{lrrr}
\toprule
& Train & Development & Test \\
\midrule
Utterances & 3,696 & 400 & 192 \\
Frames (input) & 988,012 & 107,620 & 50,380 \\
Frames (after cnn1) & 493,983 & 53,821 & 25,205 \\
Frames (after cnn2) & 233,916 & 25,469 & 11,894 \\ 
\bottomrule
\end{tabular}
\end{table}

\section{Results}

\begin{figure}[t]
\begin{subfigure}[t]{0.45\textwidth}
\includegraphics[width=\linewidth]{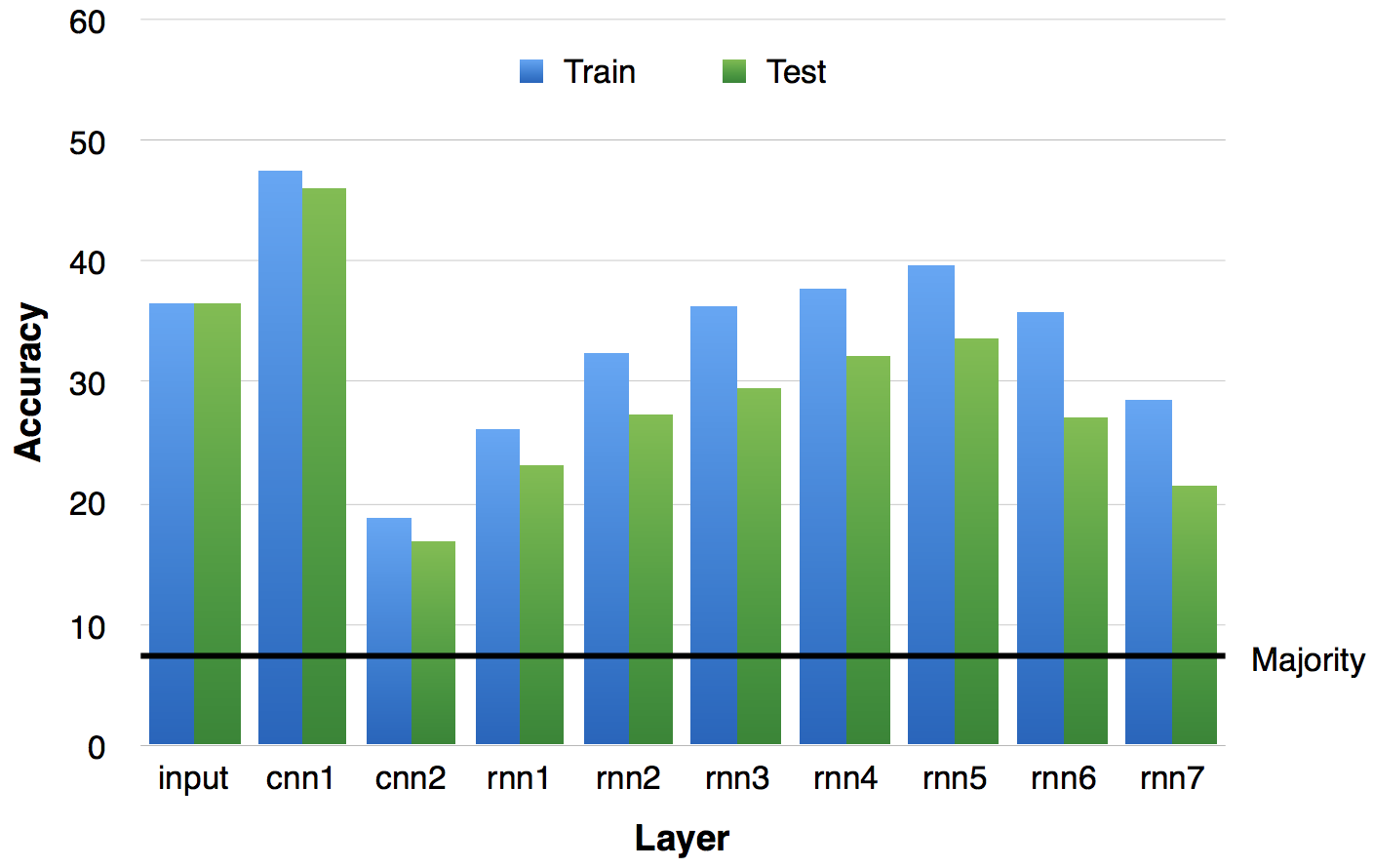}
\caption{DS2, w/ strides.}
\label{fig:results-frame}
\end{subfigure} 
\hfill
\begin{subfigure}[t]{0.45\textwidth}
\includegraphics[width=\linewidth]{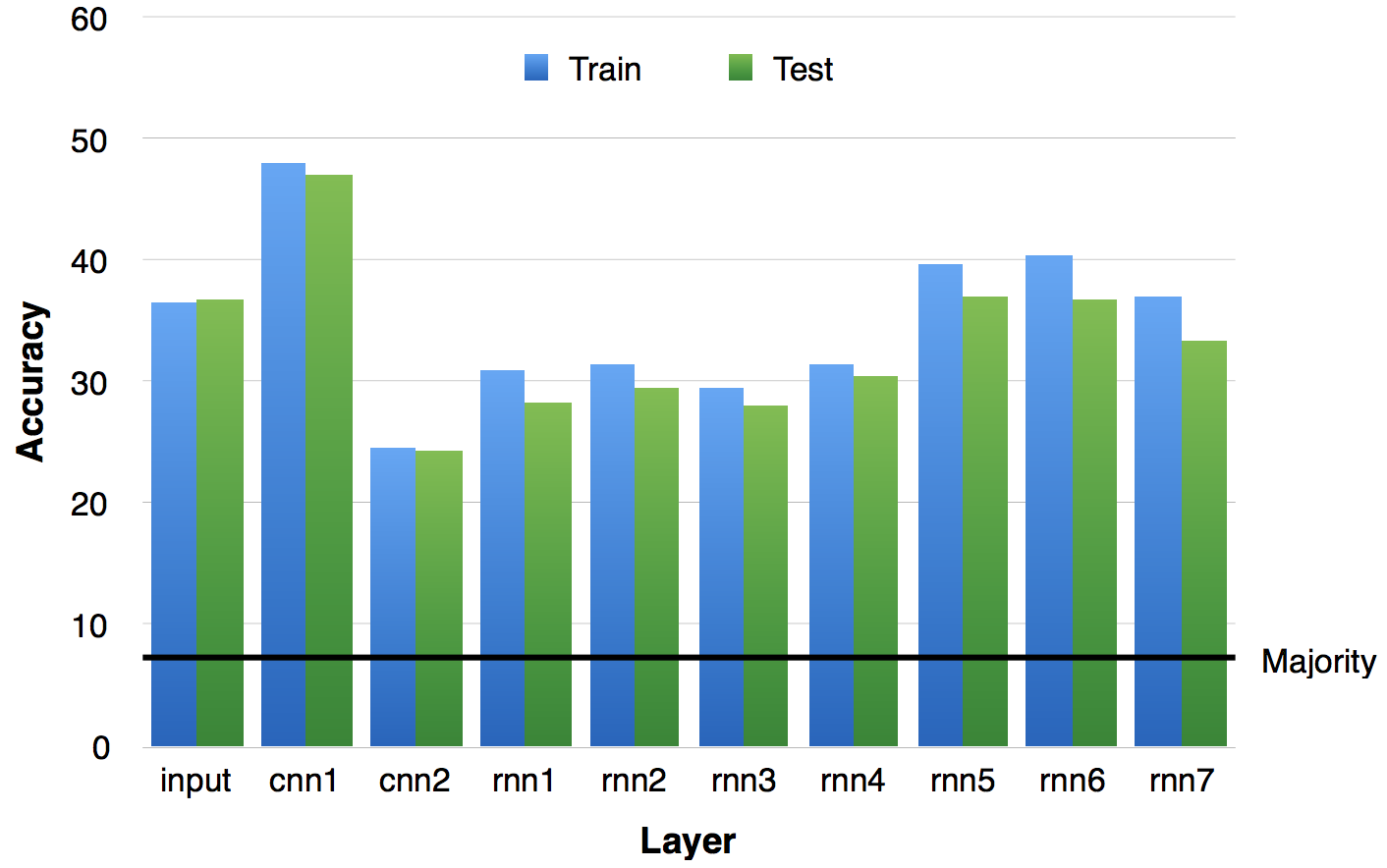}
\caption{DS2, w/o strides.}
\label{fig:results-frame-nostep}
\end{subfigure}
\hfill
\begin{subfigure}[t]{0.45\textwidth}
\includegraphics[width=\linewidth]{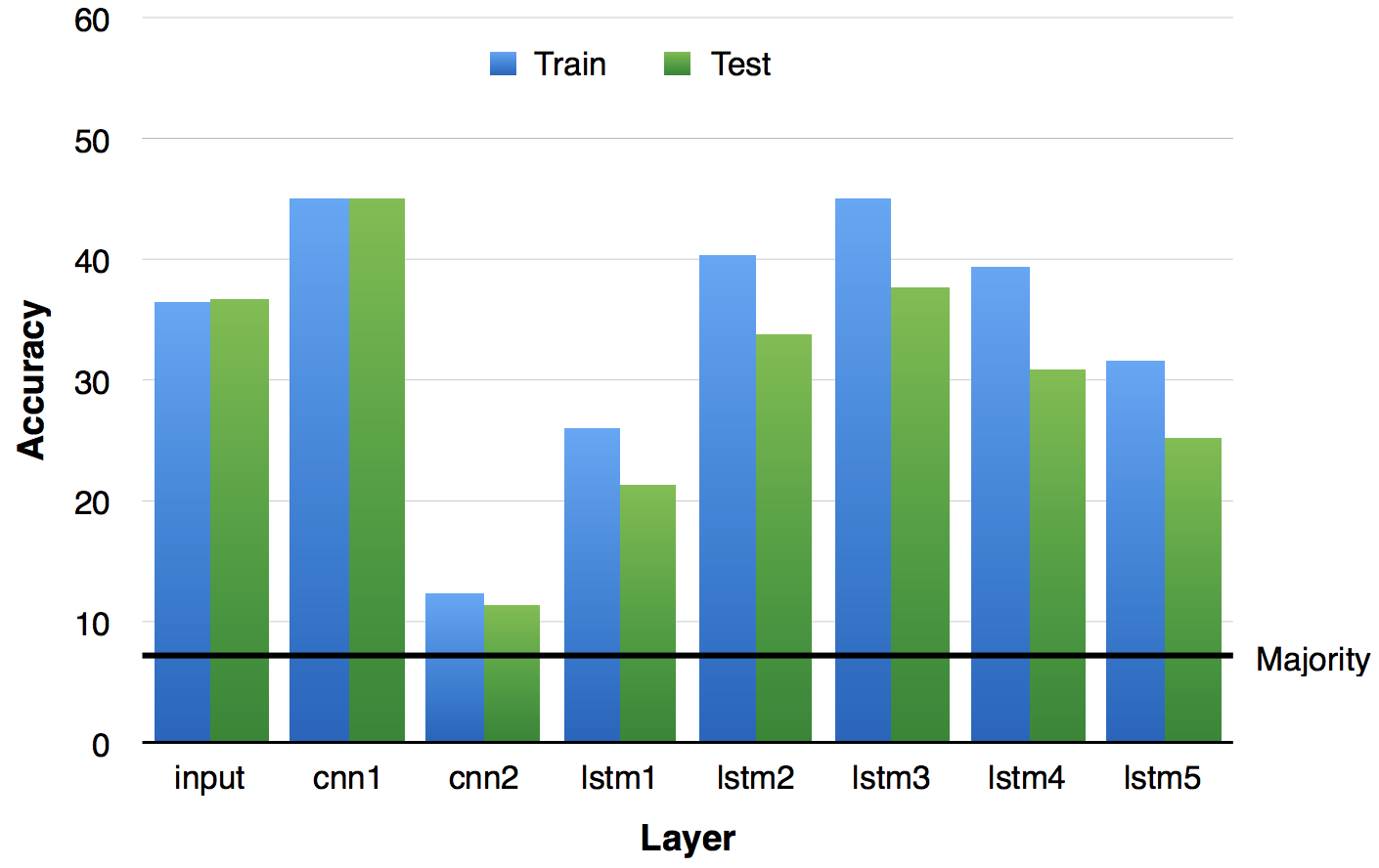}
\caption{DS2-light, w/ strides.}
\label{fig:results-frame-light}
\end{subfigure}
\hfill
\begin{subfigure}[t]{0.45\textwidth}
\includegraphics[width=\linewidth]{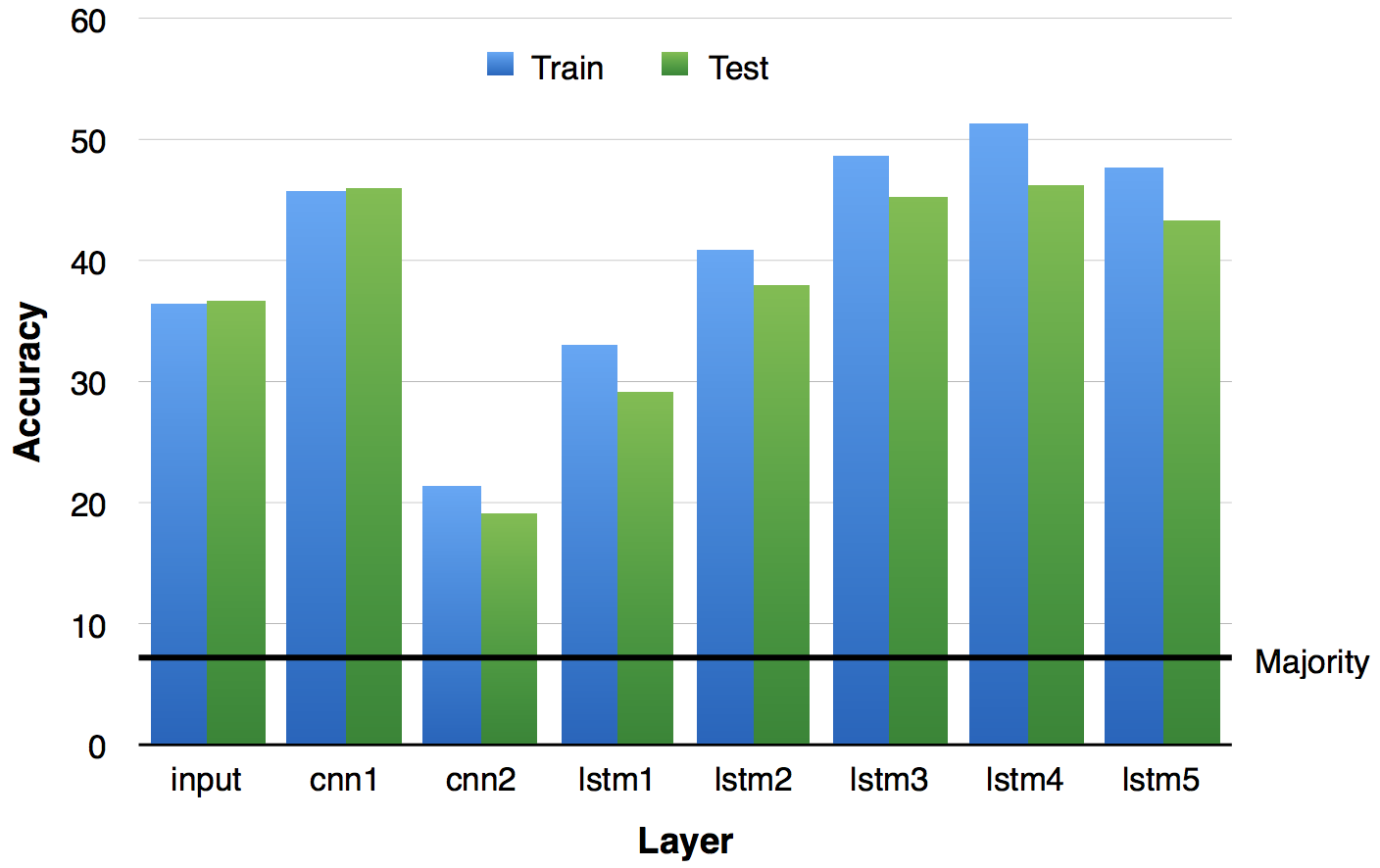}
\caption{DS2-light, w/o strides.}
\label{fig:results-frame-light-nostep}
\end{subfigure}
\caption{Frame classification accuracy using representations from different layers of DeepSpeech2 (DS2) and DeepSpeech2-light (DS2-light), with or without strides in the convolutional layers.}
\label{fig:results-frame-yesnostep}
\vspace{-2pt}
\end{figure}

Figure~\ref{fig:results-frame} shows frame classification accuracy using features from different layers of the DeepSpeech2 model. The results are all above a majority baseline of 7.25\% (the phone ``s''). Input features (spectrograms) lead to fairly good performance, considering the 60-wise classification task. The first convolution further improves the results, in line with previous findings about convolutions as feature extractors before recurrent layers~\cite{sainath2015convolutional}. However, applying a second convolution significantly degrades accuracy. This can be attributed to the filter width and stride, which may extend across phone boundaries. Nevertheless, we find the large drop quite surprising. 

The first few recurrent layers improve the results, but after the 5th recurrent layer accuracy goes down again. One possible explanation to this may be that higher layers in the model are more sensitive to long distance information that is needed for the speech recognition task, whereas  the local information that is needed for classifying phones is better captured in lower layers. For instance, to predict a word like ``bought'', the model would need to model relations between different characters, which would be better captured at the top layers. 
In contrast, feed-forward neural networks trained on phoneme recognition were shown to learn increasingly better representations at higher layers \cite{nagamine2015exploring,Nagamine+2016}; such networks do not need to model the full speech recognition task, different from end-to-end models.   

In the following sections, we first investigate three aspects of the model: model complexity, effect of strides in the convolutional layers, and effect of blanks. Then we visualize frame representations in 2D and consider classification into abstract sound classes. 
Finally, Appendix \ref{appendix-expr} provides additional experiments with windows of input features and a reduced phone set, all exhibiting similar trends. 

\subsection{Model complexity}
Figure~\ref{fig:results-frame-light} shows the results of using features from the DeepSpeech2-light model. This model has less recurrent layers (5 vs.\ 7) and smaller hidden states (600 vs.\ 1760), but it uses LSTMs instead of simple RNNs. A first observation is that the overall trend is the same as in DeepSpeech2: significant drop after the first 
convolutional layer, then initial increase followed by a drop %in accuracy 
in the final %recurrent 
layers. 

Comparing the two models (figures \ref{fig:results-frame} and \ref{fig:results-frame-light}),  a number of additional observations can be made. First, the convolutional layers of DeepSpeech2 contain more phonetic information than those of DeepSpeech2-light 
(+1\% and +4\% for cnn1 and cnn2, respectively). 
In contrast, the recurrent layers in DeepSpeech2-light are better, 
with the best result  of  37.77\% in DeepSpeech2-light (by lstm3) compared to 33.67\% in DeepSpeech2 (by rnn5).  This suggests again that 
higher layers do not model phonology very well; when there are more recurrent layers, the convolutional layers compensate and generate better representations for phonology than %in the case 
when there are fewer recurrent layers.
Interestingly, the deeper model performs better on the speech recognition task while its deep representations are not as good at capturing phonology, suggesting that its top layers focus more on modeling character sequences, while its lower layers focus on representing phonetic information.

\subsection{Effect of strides}
The original DeepSpeech2 models have convolutions with strides (steps) in the time dimension~\cite{amodei2016deep}. This leads to subsampling by a factor of 2 at each convolutional layer, resulting in reduced dataset size (Table \ref{tab:data-frame}). Consequently, the comparison between layers before and after convolutions is not entirely fair. To investigate this effect, we ran the trained convolutions without strides during feature generation for the classifier. 

Figure~\ref{fig:results-frame-nostep} shows the results %of extracting features from the DeepSpeech2 model 
at different layers without using strides in the convolutions.  The general trend is similar to the strided case: large drop at the 2nd convolutional layer, then steady increase in the recurrent layers with a drop at the final layers. However, the overall shape of the accuracy in the recurrent layers is less spiky; the initial drop is milder and performance  does not degrade as much at the top layers. A similar pattern is observed in the non-strided case of DeepSpeech2-light (Figure~\ref{fig:results-frame-light-nostep}). 

These results can be attributed to two factors. First, running convolutions without strides maintains the number of examples available to the classifier, which means a larger training set. More importantly, however, the time resolution remains high which can be important for frame classification.

\subsection{Effect of blank symbols}

Recall that the CTC model predicts either a letter in the alphabet, a space, or a blank symbol. This allows the model to concentrate probability mass on a few frames that are aligned to the output symbols in a series of spikes, separated by blank predictions~\cite{graves2006connectionist}. To investigate the effect of blank symbols on phonetic representation, we generate predictions of all symbols using the CTC model, including blanks and repetitions. Then we break down the classifier's performance into cases where the model predicted a blank, a space, or another letter. 

\begin{figure}[h]
\centering
\includegraphics[width=0.47\textwidth]{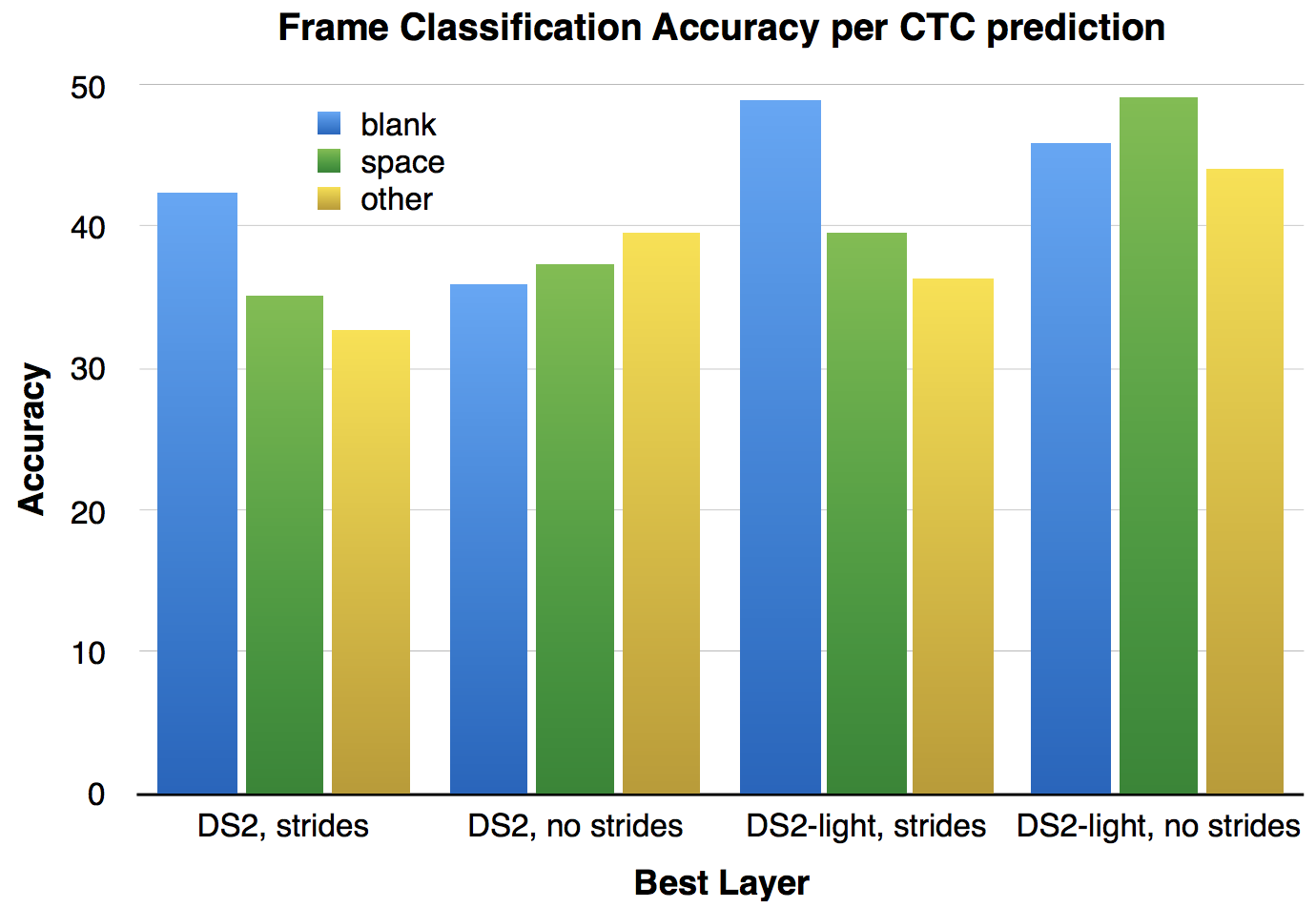}
\caption{Frame classification accuracy 
at frames predicted as blank, space, or another letter by DeepSpeech2 %(DS2) 
and DeepSpeech2-light, % (DS2-light), 
with and without strides in 
the 
convolutional layers.
}
\label{fig:results-by-ctc}
\end{figure}

Figure~\ref{fig:results-by-ctc} shows the results using representations from the best recurrent layers in DeepSpeech2 and DeepSpeech2-light, run with and without strides in the convolutional layers. In the strided case, the hidden representations are of highest quality for phone classification when the model predicts a blank. This appears counterintuitive, considering the spiky behavior of CTC models, which should be more confident when predicting non-blank. However, we found that only 5\% of the frames are predicted as blanks, due to downsampling in the strided convolutions. 
When the model is run without strides, we observe a somewhat different behavior. Note that in this case the model predicts many more blanks (more than 50\% compared to 5\% in the non-strided case),
and representations of frames predicted as blanks are not as good, which is more in line with the common spiky behavior of CTC models \cite{graves2006connectionist}. 

%%%%%% CLUSTERING %%%%%%%%

\subsection{Clustering and visualizing representations}
In this section, we visualize frame representations from different layers of DeepSpeech2. We first ran the DeepSpeech2 model on the entire development set of TIMIT and extracted feature representations for every frame from all layers. This results in more than 100K vectors of different sizes (we use the model without strides in convolutional layers to allow for comparable analysis across layers). We followed a similar procedure to that of~\cite{harwath2017learning}: We clustered the vectors in each layer with k-means ($k=500$) and plotted the cluster centroids using t-SNE~\cite{maaten2008visualizing}. We assigned to each cluster the phone label that had the largest number of examples in the cluster. As some clusters are quite noisy, we also consider pruning clusters where the majority label does not cover enough of the cluster members.

\begin{figure*}[h]
\begin{subfigure}{0.33\textwidth}
\includegraphics[width=\textwidth]{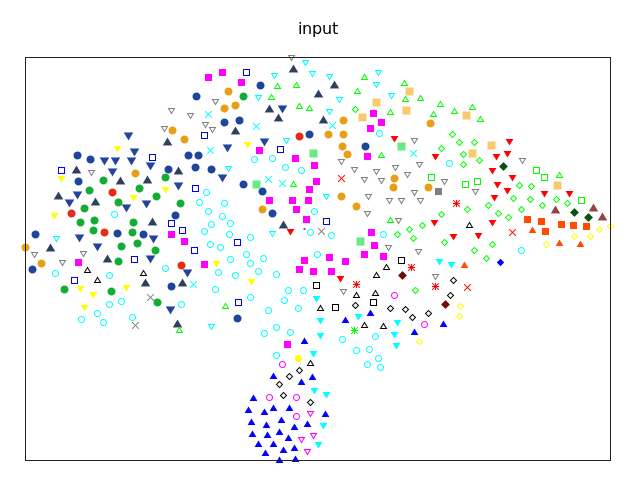}
\end{subfigure}\hfill%
\begin{subfigure}{0.33\textwidth}
\includegraphics[width=\textwidth]{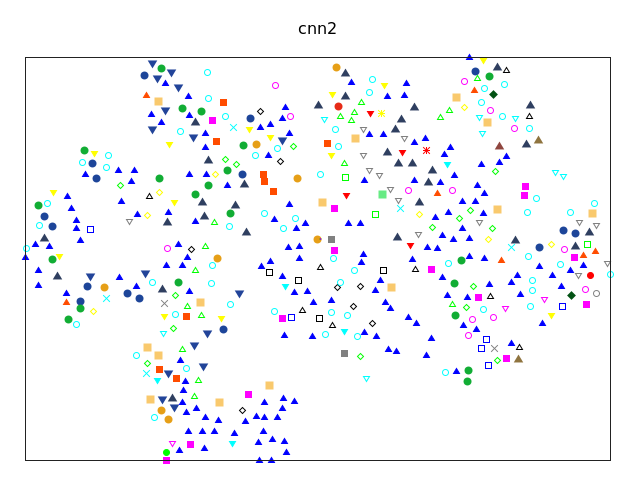}
\end{subfigure}\hfill%
\begin{subfigure}{0.33\textwidth}
\includegraphics[width=\textwidth]{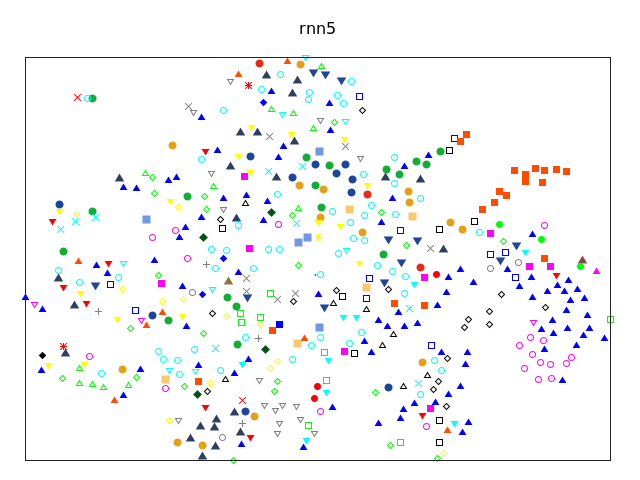}
\end{subfigure}\hfill%
\begin{subfigure}{\textwidth}
\includegraphics[width=\textwidth]{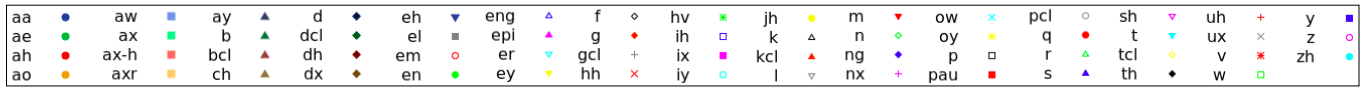}
\end{subfigure}
\caption{Centroids of frame representation clusters using features from different layers.}
\label{fig:clusters-selected}
\end{figure*}

Figure~\ref{fig:clusters-selected} shows t-SNE plots of cluster centroids from selected layers, with color and shape coding for the phone labels (see Figure~\ref{fig:clusters} in Appendix \ref{appendix-clusters} for other layers). The input layer produces clusters which show a fairly clean separation into groups of centroids with the same assigned phone. After the input layer it is less easy to detect groups, and lower layers do not show a clear structure. In layers rnn4 and rnn5 we again see some meaningful groupings (e.g.\ ``z'' on the right side of the rnn5 plot), 
after which rnn6 and rnn7 again show less structure. 

\newcommand{\myheight}{2.5cm}

Figure~\ref{fig:clusters-threshold} (in Appendix \ref{appendix-clusters}) shows clusters that have a majority label of at least 10-20\% of the examples (depending on the number of examples left in each cluster after pruning).  In this case groupings are more observable in all layers, and especially in layer rnn5. 

We note that these observations are mostly in line with our previous findings regarding the quality of representations from different layers. %It appears that 
When frame  representations are better separated in vector space, the classifier does a better job at classifying frames into their phone labels; see also \cite{Nagamine+2016}.

\subsection{Sound classes} \label{sec:sound-classes}
Speech sounds are often organized in coarse categories like consonants and vowels. In this section, 
we investigate whether the ASR model learns such categories. The primary question we ask is: which parts of the model capture most information about coarse categories? Are higher layer representations more informative for this kind of abstraction above phones? To answer this, we map %each % all TIMIT 
phones to %into 
their corresponding  %one of the following 
classes: 
affricates, fricatives, nasals, semivowels/glides, stops, and vowels. %\footnote{We also add two separate classes for pause (``pau'') and epenthetic silence (``epi''), which are annotated in the dataset and do not fall into the other classes.} 
Then we train classifiers to predict sound classes given representations from different layers of the ASR model. 

\begin{figure}[h]
\centering
\begin{minipage}{0.45\textwidth}
\includegraphics[width=0.9\linewidth]{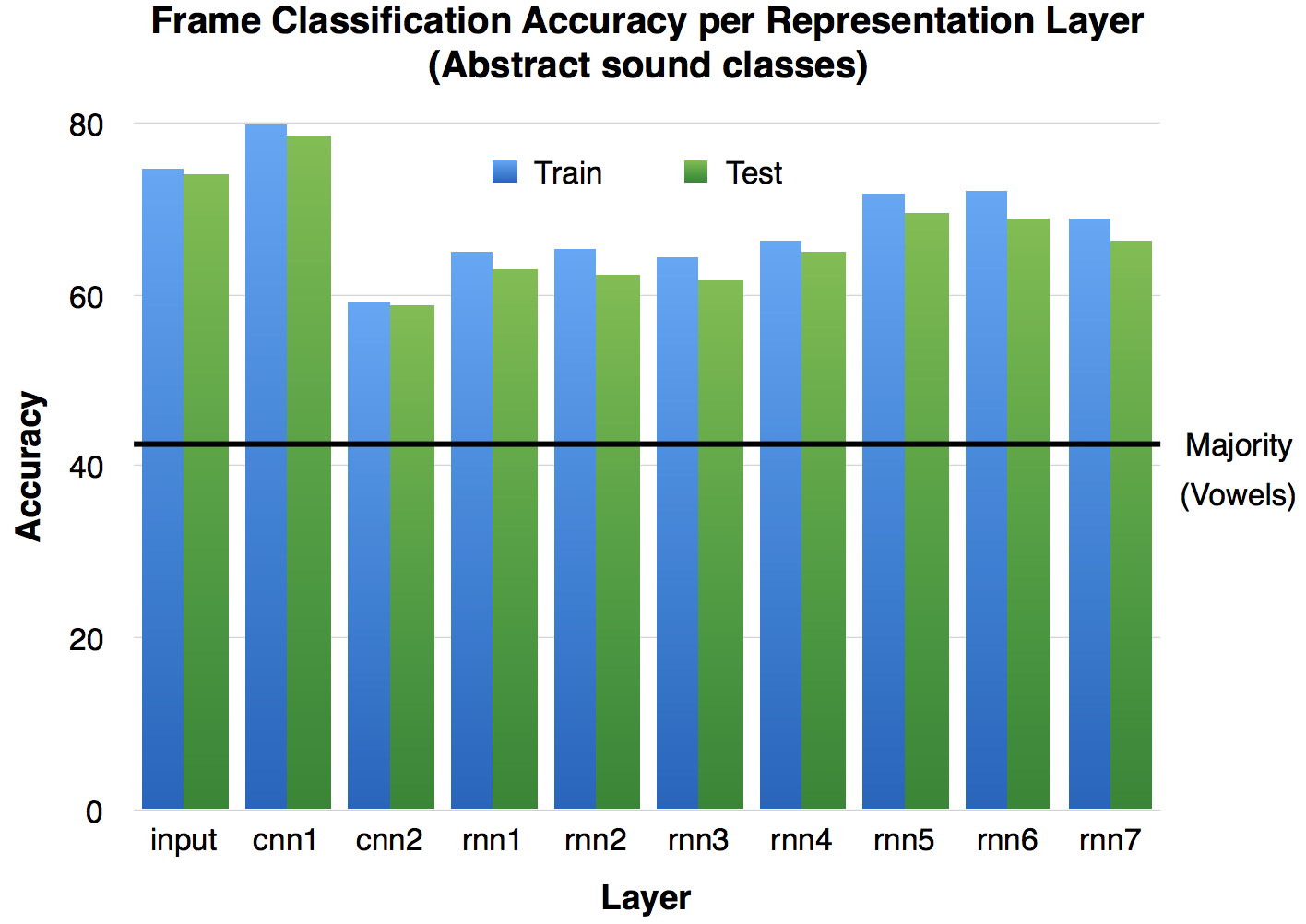}
\caption{Accuracy of classification into sound classes using representations from different layers of DeepSpeech2.}
\label{fig:results-frame-sound-class}
\end{minipage}\hfill
\begin{minipage}{0.45\textwidth}
\centering
\includegraphics[width=0.9\linewidth]{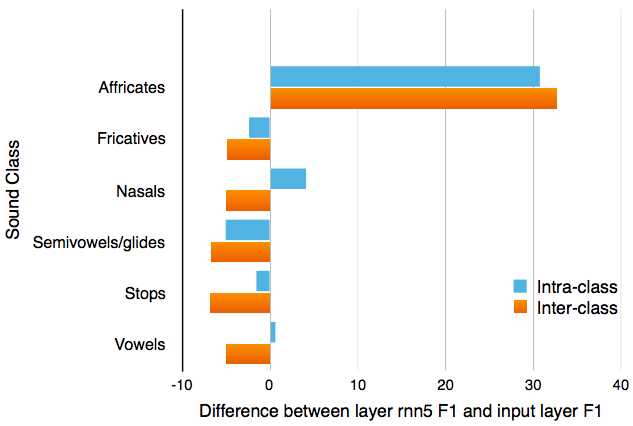}
\caption{Difference in F1 score using representations from layer rnn5 compared to the input layer. %, showing F1 within each sound class (``intra-class'') and among different classes (``inter-class'').
}
\label{fig:results-sound-class-f1-diff}
\end{minipage}
\end{figure}

Figure~\ref{fig:results-frame-sound-class} shows the results. All layers produce representations that contain a non-trivial amount of information about sound classes (above the vowel majority baseline). As expected, predicting sound classes is easier than predicting phones, as evidenced by a much higher accuracy compared to our previous results. 
As in previous experiments, the lower layers of the network (input and cnn1) produce the best representations for predicting sound classes. Performance then first drops at cnn2 and increases steadily with each recurrent layer, finally decreasing at the last recurrent layer. It appears that higher layers do not generate better representations for abstract sound classes.

Next we analyze the difference between the input layer and the best recurrent layer (rnn5), broken down to specific sound classes. We calculate the change in F1 score when moving from input representations to rnn5 representations, where F1 is calculated in two ways.  The \textit{inter-class} F1 is calculated by directly predicting coarse sound classes, thus measuring how often the model confuses two separate sound classes.
The \textit{intra-class} F1 is obtained by predicting fine-grained phones and micro-averaging F1 inside each coarse sound class (not counting confusion outside the class). It indicates how often the model confuses different phones in the same sound class.

As Figure~\ref{fig:results-sound-class-f1-diff} shows, in most cases %sound classes 
representations from rnn5 degrade the performance, both within and across classes. There are two notable exceptions. Affricates are better predicted at the higher layer, both compared to other sound classes and when predicting individual affricates. 
It may be that more contextual information is needed in order to detect a complex sound like an affricate. 
Second, the intra-class F1 for nasals improves with representations from rnn5, whereas the inter-class F1 goes down, suggesting that rnn5 is better at distinguishing between different nasals.

\begin{figure}[t]
\begin{subfigure}{0.32\textwidth}
\includegraphics[width=\linewidth]{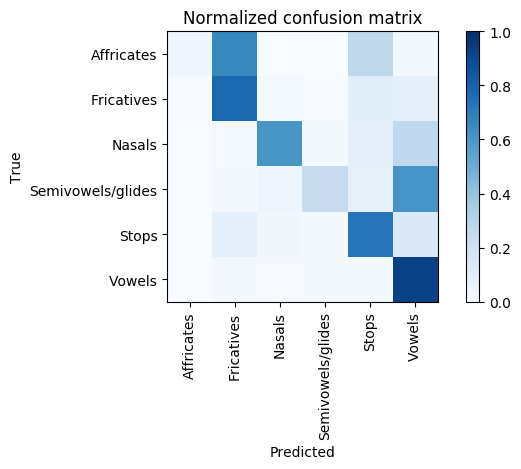}
\caption{input}
\label{fig:cm-classes-input}
\end{subfigure}
\hfill
\begin{subfigure}{0.32\textwidth}
\includegraphics[width=\linewidth]{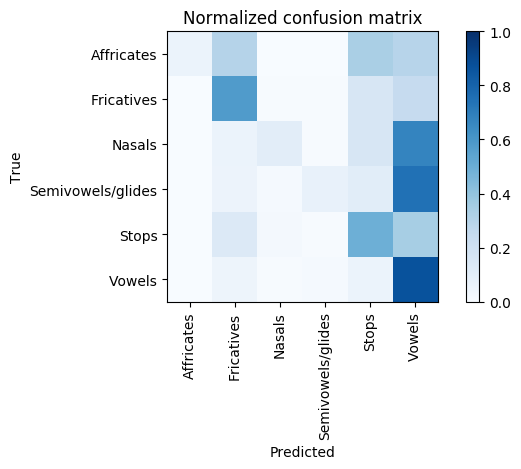}
\caption{cnn2}
\label{fig:cm-classes-cnn}
\end{subfigure}
\hfill
\begin{subfigure}{0.32\textwidth}
\includegraphics[width=\linewidth]{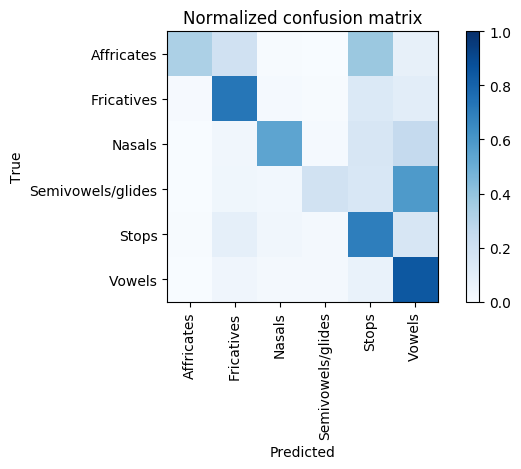}
\caption{rnn5}
\label{fig:cm-classes-rnn5}
\end{subfigure}
\caption{Confusion matrices of sound class classification %when predicting sound classes using 
using representations from different layers.
}
\label{fig:cm-classes}
%\vspace{-10pt}
\end{figure}

Finally, Figure~\ref{fig:cm-classes} shows confusion matrices of predicting sound classes using representations from the input, cnn2, and rnn5 layers. Much of the confusion arises from confusing relatively similar classes: semivowels/vowels, affricates/stops, affricates/fricatives. 
Interestingly, affricates are less confused at layer rnn5 than in lower layers, which is consistent with our previous observation.

\section{Conclusion}
In this work, we analyzed representations in a deep end-to-end ASR model that is trained with a CTC loss. We empirically evaluated the quality of the representations on a frame classification task, where each frame is classified into its corresponding phone label. We compared feature representations from different layers of the ASR model and observed striking differences in their quality. We also found that these differences are partly correlated with the separability of the representations in vector space. 

In future work, we would like to extend this analysis to other speech features, such as speaker and dialect ID, and to larger speech recognition datasets. We are also interested in experimenting with other end-to-end systems, such as sequence-to-sequence models and acoustics-to-words systems. 
Another venue for future work is to improve the end-to-end model based on our insights, for example by improving the representation capacity of certain layers in the deep neural network. 

\section*{Acknowledgements}
We would like to thank members of the MIT spoken language systems group for helpful discussions. 
This work was supported by the Qatar Computing Research Institute (QCRI). 
Any opinions, findings, conclusions, or recommendations expressed in this paper are those of the authors, and do not necessarily reflect the views of the funding organizations.

\bibliographystyle{IEEEtran}
\bibliography{mybib}

\begin{thebibliography}{10}
\providecommand{\url}[1]{#1}
\csname url@samestyle\endcsname
\providecommand{\newblock}{\relax}
\providecommand{\bibinfo}[2]{#2}
\providecommand{\BIBentrySTDinterwordspacing}{\spaceskip=0pt\relax}
\providecommand{\BIBentryALTinterwordstretchfactor}{4}
\providecommand{\BIBentryALTinterwordspacing}{\spaceskip=\fontdimen2\font plus
\BIBentryALTinterwordstretchfactor\fontdimen3\font minus
  \fontdimen4\font\relax}
\providecommand{\BIBforeignlanguage}[2]{{%
\expandafter\ifx\csname l@#1\endcsname\relax
\typeout{** WARNING: IEEEtran.bst: No hyphenation pattern has been}%
\typeout{** loaded for the language `#1'. Using the pattern for}%
\typeout{** the default language instead.}%
\else
\language=\csname l@#1\endcsname
\fi
#2}}
\providecommand{\BIBdecl}{\relax}
\BIBdecl

\bibitem{icml2014c2_graves14}
A.~Graves and N.~Jaitly, ``{Towards End-To-End Speech Recognition with
  Recurrent Neural Networks},'' in \emph{Proceedings of the 31st International
  Conference on Machine Learning (ICML-14)}, T.~Jebara and E.~P. Xing,
  Eds.\hskip 1em plus 0.5em minus 0.4em\relax JMLR Workshop and Conference
  Proceedings, 2014, pp. 1764--1772.

\bibitem{miao2015eesen}
Y.~Miao, M.~Gowayyed, and F.~Metze, ``{EESEN: End-to-end speech recognition
  using deep RNN models and WFST-based decoding},'' in \emph{2015 IEEE Workshop
  on Automatic Speech Recognition and Understanding (ASRU)}.\hskip 1em plus
  0.5em minus 0.4em\relax IEEE, 2015, pp. 167--174.

\bibitem{chorowski2014end}
J.~Chorowski, D.~Bahdanau, K.~Cho, and Y.~Bengio, ``{End-to-end Continuous
  Speech Recognition using Attention-based Recurrent NN: First Results},''
  \emph{arXiv preprint arXiv:1412.1602}, 2014.

\bibitem{chan2015listen}
W.~Chan, N.~Jaitly, Q.~V. Le, and O.~Vinyals, ``{Listen, Attend and Spell: A
  Neural Network for Large Vocabulary Conversational Speech Recognition},'' in
  \emph{2016 IEEE International Conference on Acoustics, Speech and Signal
  Processing (ICASSP)}.\hskip 1em plus 0.5em minus 0.4em\relax IEEE, 2016, pp.
  4960--4964.

\bibitem{toshniwal2017multitask}
S.~Toshniwal, H.~Tang, L.~Lu, and K.~Livescu, ``{Multitask Learning with
  Low-Level Auxiliary Tasks for Encoder-Decoder Based Speech Recognition},''
  \emph{arXiv preprint arXiv:1704.01631}, 2017.

\bibitem{Eyben:2009}
F.~Eyben, M.~Wöllmer, B.~Schuller, and A.~Graves, ``{From Speech to Letters -
  Using a Novel Neural Network Architecture for Grapheme Based ASR},'' in
  \emph{2009 IEEE Workshop on Automatic Speech Recognition and Understanding
  (ASRU)}, Nov 2009, pp. 376--380.

\bibitem{amodei2016deep}
D.~Amodei, R.~Anubhai, E.~Battenberg, C.~Case, J.~Casper, B.~Catanzaro,
  J.~Chen, M.~Chrzanowski, A.~Coates, G.~Diamos \emph{et~al.}, ``{Deep Speech
  2: End-to-End Speech Recognition in English and Mandarin},'' in
  \emph{Proceedings of The 33rd International Conference on Machine Learning},
  2016, pp. 173--182.

\bibitem{bahdanau2016end}
D.~Bahdanau, J.~Chorowski, D.~Serdyuk, P.~Brakel, and Y.~Bengio, ``{End-to-End
  Attention-based Large Vocabulary Speech Recognition},'' in \emph{2016 IEEE
  International Conference on Acoustics, Speech and Signal Processing
  (ICASSP)}.\hskip 1em plus 0.5em minus 0.4em\relax IEEE, 2016, pp. 4945--4949.

\bibitem{soltau2016neural}
H.~Soltau, H.~Liao, and H.~Sak, ``{Neural Speech Recognizer: Acoustic-to-Word
  LSTM Model for Large Vocabulary Speech Recognition},'' in \emph{Interspeech
  2017}, 2017.

\bibitem{audhkhasi2017direct}
K.~Audhkhasi, B.~Ramabhadran, G.~Saon, M.~Picheny, and D.~Nahamoo, ``{Direct
  Acoustics-to-Word Models for English Conversational Speech Recognition},'' in
  \emph{Interspeech 2017}, 2017.

\bibitem{mohamed2012understanding}
A.-r. Mohamed, G.~Hinton, and G.~Penn, ``Understanding how deep belief networks
  perform acoustic modelling,'' in \emph{2012 IEEE International Conference on
  Acoustics, Speech and Signal Processing (ICASSP)}.\hskip 1em plus 0.5em minus
  0.4em\relax IEEE, 2012, pp. 4273--4276.

\bibitem{yu2013feature}
D.~Yu, M.~L. Seltzer, J.~Li, J.-T. Huang, and F.~Seide, ``{Feature Learning in
  Deep Neural Networks - Studies on Speech Recognition Tasks},'' in
  \emph{International Conference on Learning Representations (ICLR)}, 2013.

\bibitem{nagamine2015exploring}
T.~Nagamine, M.~L. Seltzer, and N.~Mesgarani, ``{Exploring How Deep Neural
  Networks Form Phonemic Categories},'' in \emph{Interspeech 2015}, 2015.

\bibitem{Nagamine+2016}
------, ``{On the Role of Nonlinear Transformations in Deep Neural Network
  Acoustic Models},'' in \emph{Interspeech 2016}, 2016, pp. 803--807.

\bibitem{wang2017gate}
Y.-H. Wang, C.-T. Chung, and H.-y. Lee, ``{Gate Activation Signal Analysis for
  Gated Recurrent Neural Networks and Its Correlation with Phoneme
  Boundaries},'' in \emph{Interspeech 2017}, 2017.

\bibitem{wu2016investigating}
Z.~Wu and S.~King, ``Investigating gated recurrent networks for speech
  synthesis,'' in \emph{2016 IEEE International Conference on Acoustics, Speech
  and Signal Processing (ICASSP)}.\hskip 1em plus 0.5em minus 0.4em\relax IEEE,
  2016, pp. 5140--5144.

\bibitem{chaabouni2017learning}
R.~Chaabouni, E.~Dunbar, N.~Zeghidour, and E.~Dupoux, ``Learning weakly
  supervised multimodal phoneme embeddings,'' in \emph{Interspeech 2017}, 2017.

\bibitem{chrupala2017representations}
G.~Chrupa{\l}a, L.~Gelderloos, and A.~Alishahi, ``{Representations of language
  in a model of visually grounded speech signal},'' in \emph{Proceedings of the
  55th Annual Meeting of the Association for Computational Linguistics (Volume
  1: Long Papers)}.\hskip 1em plus 0.5em minus 0.4em\relax Association for
  Computational Linguistics, 2017, pp. 613--622.

\bibitem{harwath2017learning}
D.~Harwath and J.~Glass, ``{Learning Word-Like Units from Joint Audio-Visual
  Analysis},'' in \emph{Proceedings of the 55th Annual Meeting of the
  Association for Computational Linguistics (Volume 1: Long Papers)}.\hskip 1em
  plus 0.5em minus 0.4em\relax Association for Computational Linguistics, 2017,
  pp. 506--517.

\bibitem{K17-1037}
A.~Alishahi, M.~Barking, and G.~Chrupa{\l}a, ``Encoding of phonology in a
  recurrent neural model of grounded speech,'' in \emph{Proceedings of the 21st
  Conference on Computational Natural Language Learning (CoNLL 2017)}.\hskip
  1em plus 0.5em minus 0.4em\relax Association for Computational Linguistics,
  2017, pp. 368--378.

\bibitem{shi-padhi-knight:2016:EMNLP2016}
X.~Shi, I.~Padhi, and K.~Knight, ``{Does String-Based Neural MT Learn Source
  Syntax?}'' in \emph{Proceedings of the 2016 Conference on Empirical Methods
  in Natural Language Processing}.\hskip 1em plus 0.5em minus 0.4em\relax
  Austin, Texas: Association for Computational Linguistics, November 2016, pp.
  1526--1534.

\bibitem{belinkov:2017:ACL}
Y.~Belinkov, N.~Durrani, F.~Dalvi, H.~Sajjad, and J.~Glass, ``{What do Neural
  Machine Translation Models Learn about Morphology?}'' in \emph{Proceedings of
  the 55th Annual Meeting of the Association for Computational Linguistics
  (Volume 1: Long Papers)}.\hskip 1em plus 0.5em minus 0.4em\relax Association
  for Computational Linguistics, 2017, pp. 861--872.

\bibitem{gelderloos-chrupala:2016:COLING}
L.~Gelderloos and G.~Chrupa{\l}a, ``From phonemes to images: levels of
  representation in a recurrent neural model of visually-grounded language
  learning,'' in \emph{Proceedings of COLING 2016, the 26th International
  Conference on Computational Linguistics: Technical Papers}.\hskip 1em plus
  0.5em minus 0.4em\relax Osaka, Japan: The COLING 2016 Organizing Committee,
  December 2016, pp. 1309--1319.

\bibitem{hochreiter1997long}
S.~Hochreiter and J.~Schmidhuber, ``{Long short-term memory},'' \emph{Neural
  Computation}, vol.~9, no.~8, pp. 1735--1780, 1997.

\bibitem{ioffe2015batch}
S.~Ioffe and C.~Szegedy, ``{Batch Normalization: Accelerating Deep Network
  Training by Reducing Internal Covariate Shift},'' in \emph{Proceedings of the
  32Nd International Conference on International Conference on Machine Learning
  (ICML)}, vol.~37, 2015, pp. 448--456.

\bibitem{laurent2016batch}
C.~Laurent, G.~Pereyra, P.~Brakel, Y.~Zhang, and Y.~Bengio, ``{Batch Normalized
  Recurrent Neural Networks},'' in \emph{2016 IEEE International Conference on
  Acoustics, Speech and Signal Processing (ICASSP)}.\hskip 1em plus 0.5em minus
  0.4em\relax IEEE, 2016, pp. 2657--2661.

\bibitem{graves2006connectionist}
A.~Graves, S.~Fern{\'a}ndez, F.~Gomez, and J.~Schmidhuber, ``{Connectionist
  Temporal Classification: Labelling Unsegmented Sequence Data with Recurrent
  Neural Networks},'' in \emph{Proceedings of the 23rd International Conference
  on Machine Learning (ICML)}, 2006, pp. 369--376.

\bibitem{kingma2014adam}
D.~Kingma and J.~Ba, ``{Adam: A Method for Stochastic Optimization},''
  \emph{arXiv preprint arXiv:1412.6980}, 2014.

\bibitem{Naren2016}
S.~Naren, ``deepspeech.torch,''
  \url{https://github.com/SeanNaren/deepspeech.torch}, 2016.

\bibitem{panayotov2015librispeech}
V.~Panayotov, G.~Chen, D.~Povey, and S.~Khudanpur, ``{Librispeech: an ASR
  corpus based on public domain audio books},'' in \emph{2015 IEEE
  International Conference on Acoustics, Speech and Signal Processing
  (ICASSP)}.\hskip 1em plus 0.5em minus 0.4em\relax IEEE, 2015, pp. 5206--5210.

\bibitem{Lee-Hon:1989}
K.-F. Lee and H.-W. Hon, ``Speaker-independent phone recognition using hidden
  markov models,'' \emph{IEEE Transactions on Acoustics, Speech, and Signal
  Processing}, vol.~37, no.~11, pp. 1641--1648, 1989.

\bibitem{sainath2015convolutional}
T.~N. Sainath, O.~Vinyals, A.~Senior, and H.~Sak, ``{Convolutional, Long
  Short-Term Memory, fully connected Deep Neural Networks},'' in \emph{2015
  IEEE International Conference on Acoustics, Speech and Signal Processing
  (ICASSP)}.\hskip 1em plus 0.5em minus 0.4em\relax IEEE, 2015, pp. 4580--4584.

\bibitem{maaten2008visualizing}
L.~v.~d. Maaten and G.~Hinton, ``{Visualizing data using t-SNE},''
  \emph{Journal of Machine Learning Research}, vol.~9, pp. 2579--2605, 2008.

\bibitem{sak2015acoustic}
H.~Sak, F.~de~Chaumont~Quitry, T.~Sainath, K.~Rao \emph{et~al.}, ``{Acoustic
  Modelling with CD-CTC-SMBR LSTM RNNs},'' in \emph{2015 IEEE Workshop on
  Automatic Speech Recognition and Understanding (ASRU)}.\hskip 1em plus 0.5em
  minus 0.4em\relax IEEE, 2015, pp. 604--609.

\end{thebibliography}

\newpage

\appendix

\section{Additional experiments} \label{appendix-expr}

\subsection{Windows of features} \label{sec:window}
\begin{figure}[h]
\includegraphics[width=\linewidth]{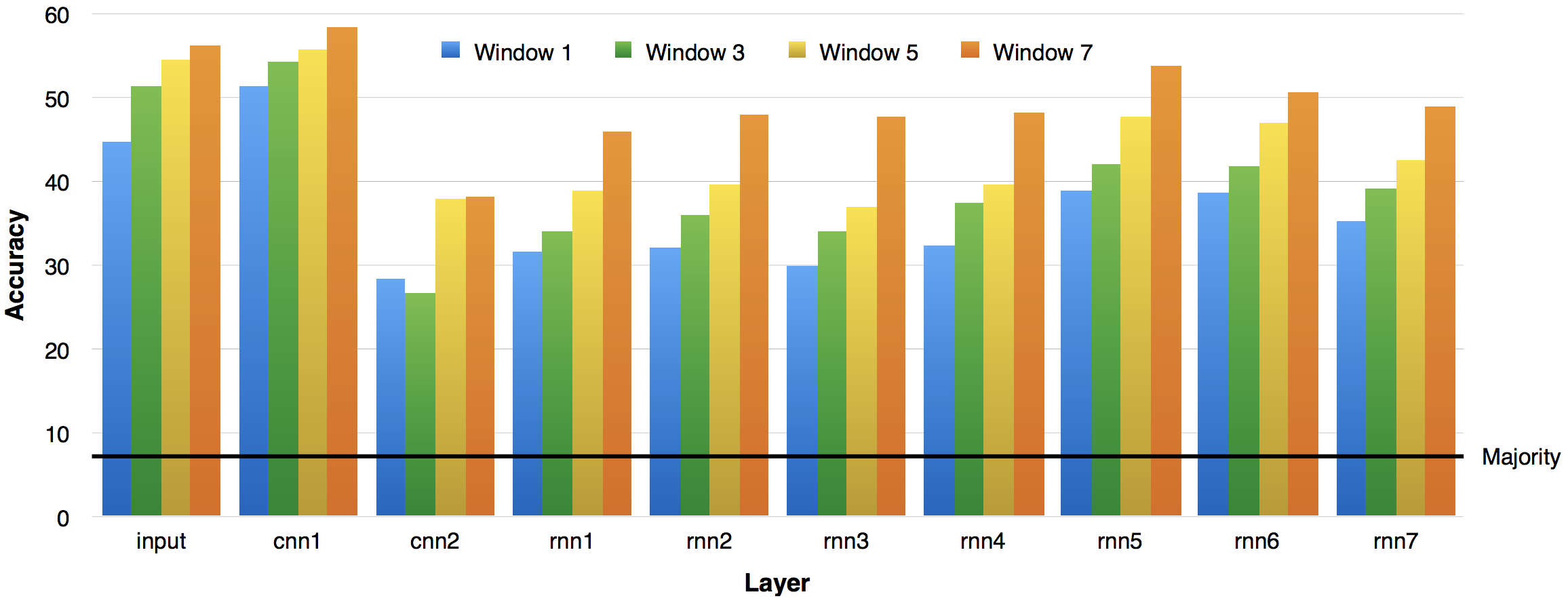}
\caption{Frame classification accuracy using different window widths around the current frame.}
\label{fig:results-window}
\end{figure}

Out main experiments used a simple frame representation by taking the output of the ASR model at frame $t$, $\texttt{ASR}_t(\xx)$. We also consider a window of features around the frame at time $t$. This improves the representation and also accounts for possible delay effects \cite{sak2015acoustic}. Test set results with different window widths are shown in Figure \ref{fig:results-window} (DeepSpeech model, no strides). 
As expected, larger windows improve the representation quality. 
The absolute numbers are much better than using only a single frame (+10-15\%), but the overall trend for a given window size is similar: initial performance drop after the convolutional layers, then steady increase at the first recurrent layers and another drop at the top layers. The drop is somewhat more moderate than in the single frame case (compare to Figure \ref{fig:results-frame-nostep}), indicating that some shifting effect may indeed be taking place, although it might be limited given that we are using bidirectional RNNs (the results in \cite{sak2015acoustic} are with unidirectional RNNs).

%\vspace{-10pt}
\subsection{Reduced phone set} \label{sec:phones-48}
\begin{figure}[h]
\centering
\includegraphics[width=0.7\linewidth]{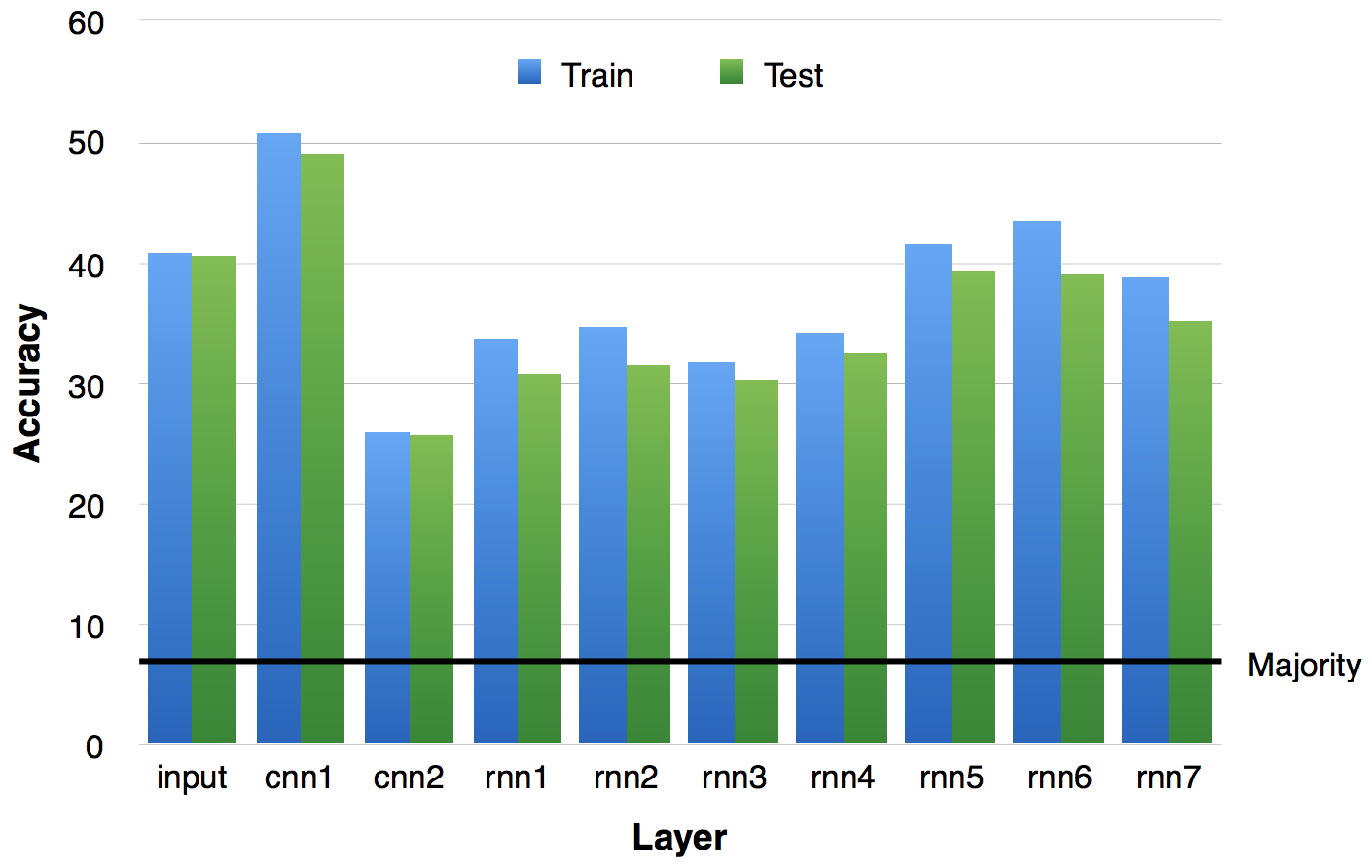}
\caption{Frame classification accuracy with a reduced set of 48 phones.}
\label{fig:results-frame-48phones}
\end{figure}

In addition to the full set of 60 phones and the coarse sound classes, we also experimented with a reduced set of 48 phones \cite{Lee-Hon:1989}. As Figure \ref{fig:results-frame-48phones} shows, the trend is similar to the other phone sets. we also noticed, as with sound classes (Section \ref{sec:sound-classes}), that the affricates /jh/ and /ch/ are better represented at rnn5 than at the input layer (not shown). /jh/ even improves further at the top layer (rnn7). 

\newpage

\section{Visualizations of frame representations} \label{appendix-clusters}

\renewcommand{\myheight}{2.59cm}

\begin{figure*}[h]
\begin{subfigure}{0.3\textwidth}
\includegraphics[height=\myheight,width=\textwidth]{figs/timit_dev_input_centroids_iter5000_nolegend.png}
\end{subfigure}\hfill%
\begin{subfigure}{0.3\textwidth}
\includegraphics[height=\myheight,width=\textwidth]{figs/timit_dev_cnn_noStep_centroids_iter5000_nolegend.png}
\end{subfigure}\hfill%
\begin{subfigure}{0.3\textwidth}
\includegraphics[height=\myheight,width=\textwidth]{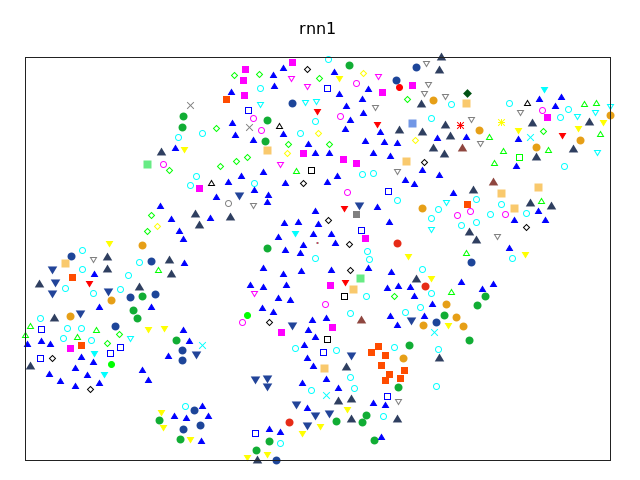}
\end{subfigure}\hfill%
\begin{subfigure}{0.3\textwidth}
\includegraphics[height=\myheight,width=\textwidth]{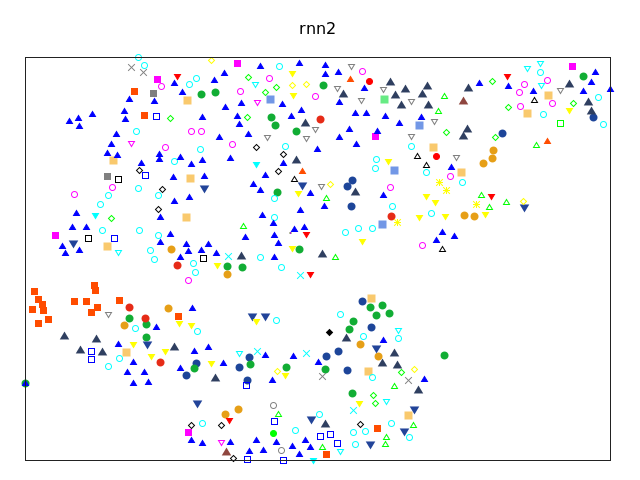}
\end{subfigure}\hfill%
\begin{subfigure}{0.3\textwidth}
\includegraphics[height=\myheight,width=\textwidth]{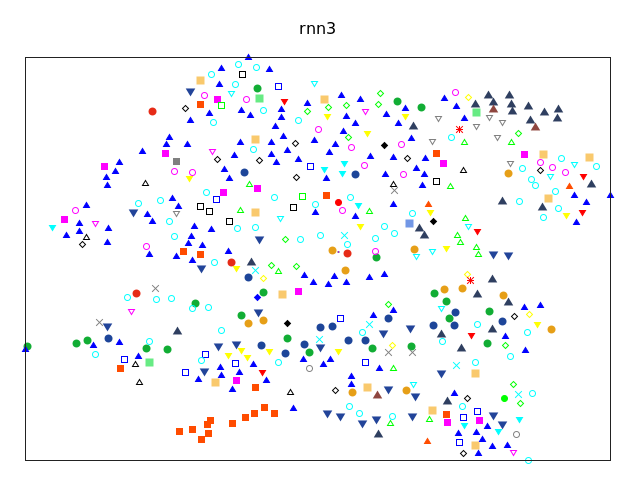}
\end{subfigure}\hfill%
\begin{subfigure}{0.3\textwidth}
\includegraphics[height=\myheight,width=\textwidth]{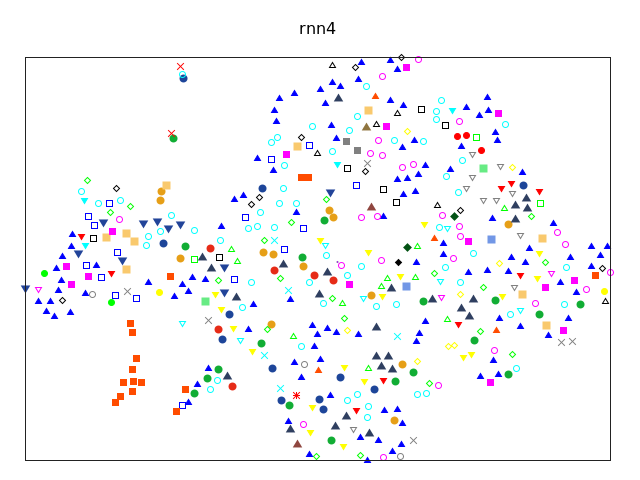}
\end{subfigure}\hfill%
\begin{subfigure}{0.3\textwidth}
\includegraphics[height=\myheight,width=\textwidth]{figs/timit_dev_rnn5_noStep_centroids_iter5000_nolegend.png}
\end{subfigure}\hfill%
\begin{subfigure}{0.3\textwidth}
\includegraphics[height=\myheight,width=\textwidth]{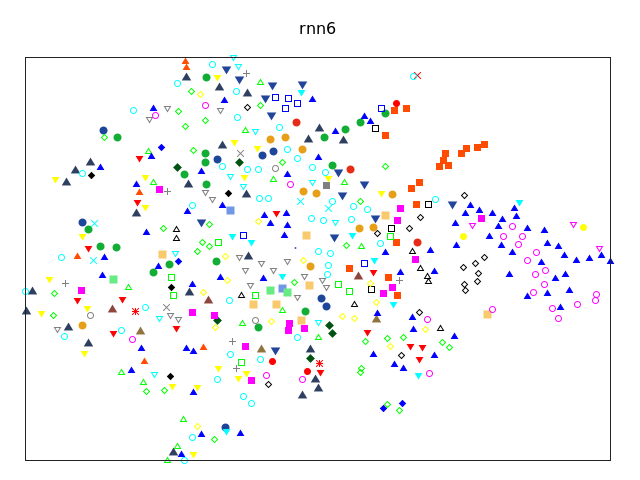}
\end{subfigure}\hfill%
\begin{subfigure}{0.3\textwidth}
\includegraphics[height=\myheight,width=\textwidth]{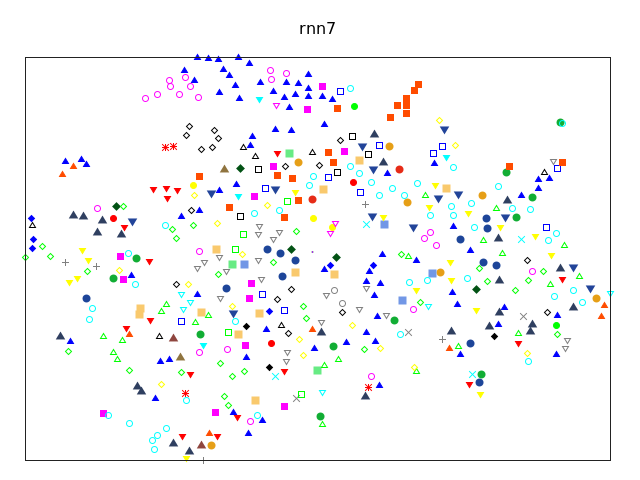}
\end{subfigure}\hfill%
\begin{subfigure}{\textwidth}
\includegraphics[width=\textwidth]{figs/legend-ppt-pdf.png}
\end{subfigure}
\caption{Centroids of all frame representation clusters using features from different layers.}
\label{fig:clusters}
\end{figure*}

\begin{figure*}[h]
\begin{subfigure}{0.3\textwidth}
\includegraphics[height=\myheight,width=\textwidth]{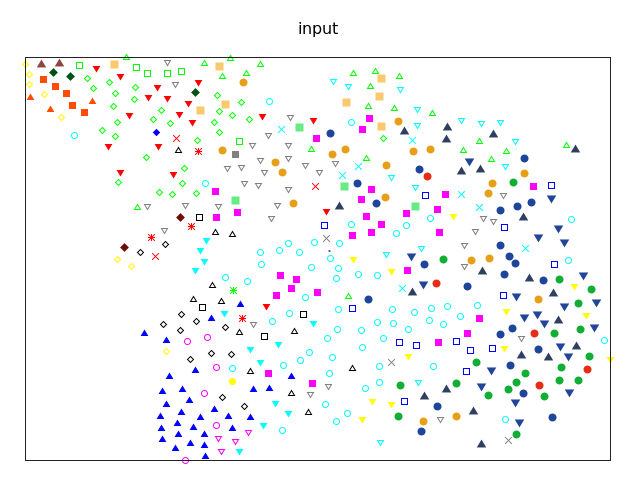}
\end{subfigure}\hfill%
\begin{subfigure}{0.3\textwidth}
\includegraphics[height=\myheight,width=\textwidth]{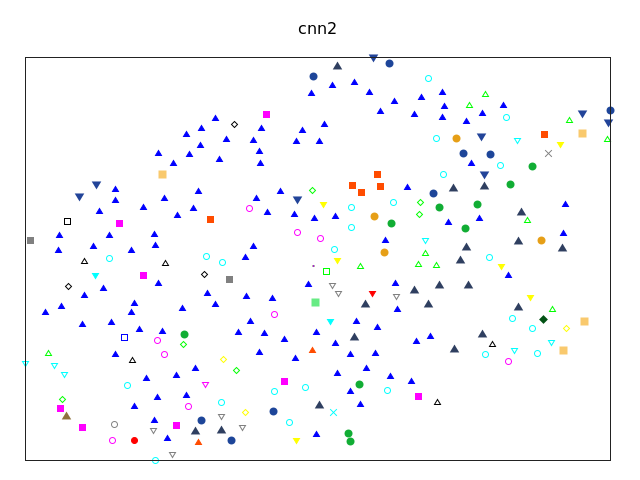}
\end{subfigure}\hfill%
\begin{subfigure}{0.3\textwidth}
\includegraphics[height=\myheight,width=\textwidth]{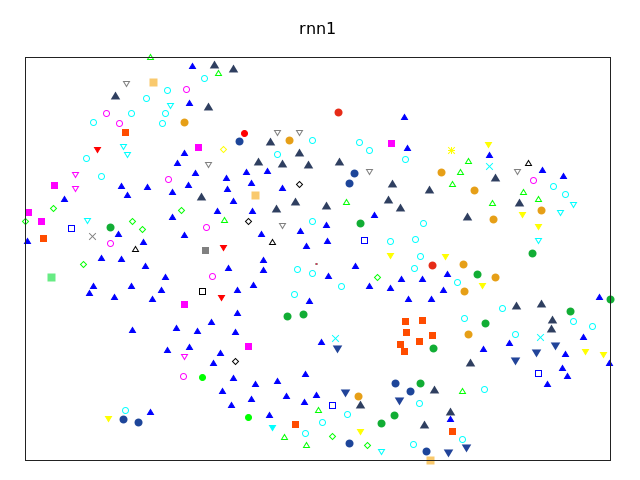}
\end{subfigure}\hfill%
\begin{subfigure}{0.3\textwidth}
\includegraphics[height=\myheight,width=\textwidth]{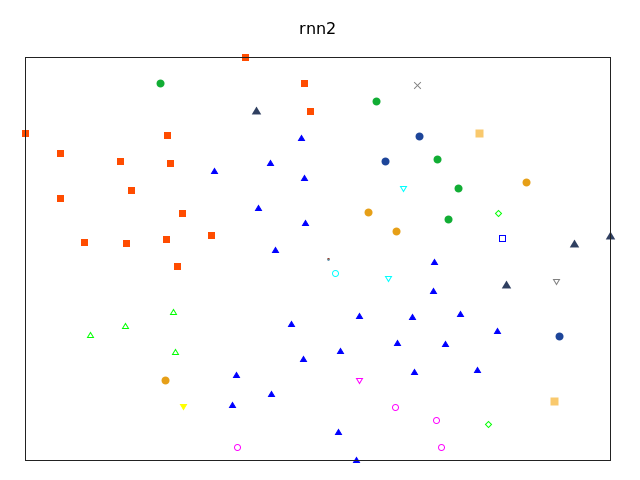}
\end{subfigure}\hfill%
\begin{subfigure}{0.3\textwidth}
\includegraphics[height=\myheight,width=\textwidth]{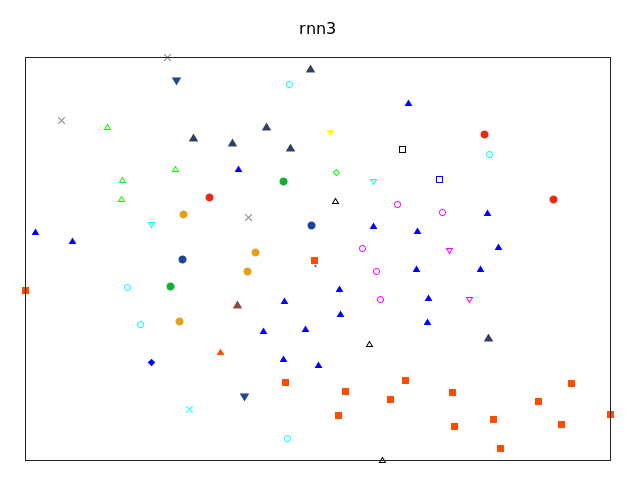}
\end{subfigure}\hfill%
\begin{subfigure}{0.3\textwidth}
\includegraphics[height=\myheight,width=\textwidth]{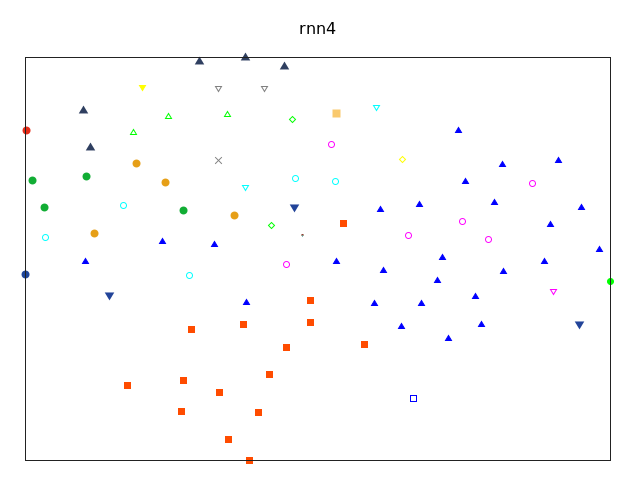}
\end{subfigure}\hfill%
\begin{subfigure}{0.3\textwidth}
\includegraphics[height=\myheight,width=\textwidth]{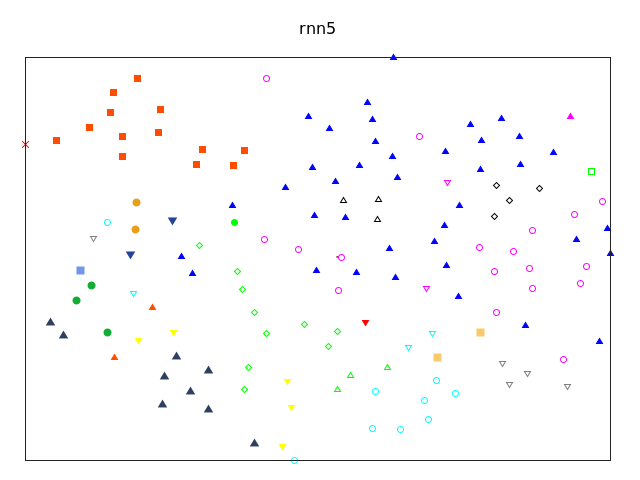}
\end{subfigure}\hfill%
\begin{subfigure}{0.3\textwidth}
\includegraphics[height=\myheight,width=\textwidth]{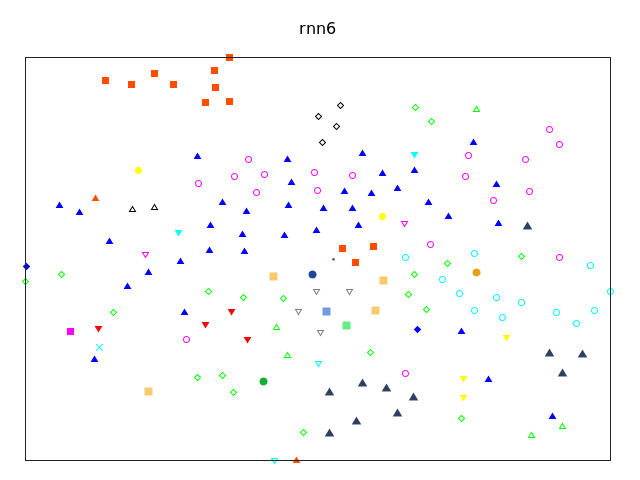}
\end{subfigure}\hfill%
\begin{subfigure}{0.3\textwidth}
\includegraphics[height=\myheight,width=\textwidth]{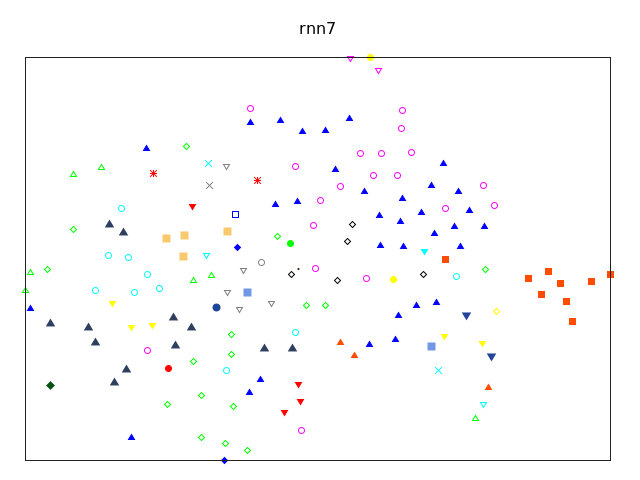}
\end{subfigure}\hfill%
\begin{subfigure}{\textwidth}
\includegraphics[width=\textwidth]{figs/legend-ppt-pdf.png}
\end{subfigure}
\caption{Centroids of frame representation clusters using features from different layers, showing only clusters where the majority label covers at least 10-20\% of the cluster members.}
\label{fig:clusters-threshold}
\end{figure*}

\end{document}